\documentclass{IEEEtran}
\IEEEoverridecommandlockouts
\usepackage{cite}
\usepackage{amsmath,amssymb,amsfonts}
\usepackage{graphicx}
\usepackage{textcomp}
\usepackage{array}
\usepackage{multirow}
\usepackage{booktabs}
\usepackage[caption=false,font=footnotesize]{subfig}
\usepackage{url}
\usepackage[colorlinks=true,urlcolor=blue,linkcolor=black,citecolor=black]{hyperref}

\newcommand{\cmIoUGC}{mIoU$_{GC}$}
\newcommand{\cmIoUG}{mIoU$_{G}$}
\newcommand{\cmIoUL}{mIoU$_{L}$}
\newcommand{\cmIoUA}{mIoU$_{A}$}

\begin{document}

\title{\LARGE GUARD: Geometric Uncertainty-Aware Point Cloud Denoising and Segmentation for Robotic Hard Disk Drive Disassembly}

\author{Zuoxu Wang, Xiao Liang$^*$
\thanks{Zuoxu Wang, Xiao Liang are with the Zachry Department of Civil and Environmental Engineering, Texas A\&M University, College Station, TX 77843 USA(e-mail: zuoxu@tamu.edu; xliang@tamu.edu).}
\thanks{$^*$ Corresponding Author.}
\thanks{This work was supported by the USA National Science Foundation under Grant No.2422826. Portions of this research were conducted with the advanced computing resources provided by
Texas A\&M High Performance Research Computing}
}

\maketitle

\begin{abstract}
Reliable robotic disassembly requires part-level representations that distinguish genuine component geometry from scanning and reconstruction artifacts. In point clouds of hard disk drives (HDDs), structured ghost artifacts can resemble valid components locally while remaining inconsistent with the overall geometry, allowing erroneous measurements to receive plausible semantic labels. This creates an engineering information problem: semantic prediction confidence alone does not establish whether the underlying geometry is reliable. 
We propose \textbf{GUARD}, a geometric uncertainty-aware framework that performs point filtering and segmentation within a single forward pass by modeling the reliability of learned geometric representations. GUARD combines a multi-scale geometric transformer with a multi-bandwidth random Fourier feature Gaussian Process to estimate per-point geometric uncertainty, complemented by predictive entropy to suppress unreliable measurements while preserving informative structures. 
Evaluation on 2,745 real HDD point clouds shows that GUARD improves PointNet++ segmentation mean intersection over union from 0.7739 to 0.8318. Additional experiments on ShapeNetPart and ScanNet examine robustness across corruption types, point-cloud domains, and segmentation backbones. On manually annotated ScanNet samples, geometric uncertainty achieves a corrupted-point detection F1 score of 0.7931, compared with 0.2212 for predictive entropy. The results demonstrate the value of distinguishing geometric reliability from semantic confidence and reveal a tradeoff between artifact suppression and preservation of informative structures. GUARD contributes a reliability-aware approach to interpreting imperfect 3D measurements for component identification and subsequent robotic handling. Project website: \href{https://001-wang.github.io/GUARD_Point_denoiser/}{\nolinkurl{https://001-wang.github.io/GUARD_Point_denoiser/}}.
\end{abstract}


\section{Introduction}
\label{sec:intro}

The rapid replacement and upgrading of electronic devices has led to a continuously increasing volume of electronic waste (e-waste), creating growing demands for efficient material recovery, component handling, and recycling. Hard disk drives (HDDs), for example, contain multiple reusable or valuable components and materials that require accurate identification, handling, sorting, and disassembly during automated processing~\cite{jin2020life, liang2023improved}. However, these operations still rely heavily on manual labor, which is time-consuming, repetitive, and difficult to scale. With recent advances in industrial robotics, robotic systems have shown increasing potential for automating component handling and disassembly, thereby improving the efficiency, consistency, and scalability of e-waste processing~\cite{zhang2024mixed, guo2023human}.

Reliable perception is a critical prerequisite for robotic disassembly~\cite{ameur2025future}. Before a robot can grasp, remove, or manipulate an individual component, it must accurately identify the corresponding part and understand its three-dimensional geometry. This requirement is particularly important for electronic devices such as HDDs, where multiple small components are densely arranged within a limited workspace~\cite{huang2025robotic}. Point clouds provide a suitable representation for this purpose because they preserve explicit 3D geometric information and allow the robot to reason about fine-grained part structures and spatial relationships. Consequently, accurate part-level point cloud segmentation is an important perception capability for automated disassembly and manipulation~\cite{ahrens2026addressing}.

In real-world deployment, however, point clouds obtained from depth sensors or multi-view reconstruction are rarely free from corruption. Sensor noise, partial observations, occlusion, reflective surfaces, and reconstruction errors can introduce inaccurate or redundant measurements~\cite{7785084,rs13163058,zhang2025deeplearningpointcloud}. More importantly, multi-frame scanning and 3D reconstruction can produce structured ghost artifacts, where duplicated, shifted, or partially reconstructed surfaces appear around the true object geometry~\cite{s25113377,10251698,jimaging9120255}. Unlike isolated floating outliers, these ghost structures may preserve locally coherent shapes that closely resemble portions of real components. As a result, they can be particularly difficult to distinguish from valid points and may cause segmentation networks to assign plausible but incorrect semantic labels.

Although point cloud denoising has been extensively studied, most existing methods primarily focus on random perturbations, sparse outliers, or general surface noise. For example, supervised denoising methods such as PointCleanNet~\cite{rakotosaona2019pointcleannet} learn to remove corrupted points from paired clean and noisy data, while more recent robustness-oriented approaches attempt to suppress high-risk or unreliable samples. However, structured ghost artifacts receive considerably less attention. This type of corruption is fundamentally more challenging because the erroneous points are not necessarily isolated or geometrically irregular; instead, they may form locally realistic structures similar to genuine object parts. Moreover, obtaining point-level annotations of such artifacts from real scans is expensive and impractical. Therefore, reliably identifying geometrically unreliable regions while preserving valid fine-grained structures remains an important challenge for robust point cloud segmentation in real robotic perception.

Motivated by these limitations, this study investigates how geometric consistency and uncertainty can be represented and used to improve part-level interpretation of artifact-corrupted engineering scans. In particular, local geometric plausibility and confident semantic prediction do not necessarily imply that a measured structure is valid. A useful perception method must therefore evaluate the reliability of geometric evidence while preserving the fine structures needed for component recognition.

We address this question through GUARD, a framework that combines multi-scale geometric representations, geometry-conditioned uncertainty estimation, and semantic prediction entropy. Rather than relying solely on classification confidence, GUARD evaluates how well a point’s geometric representation is supported by the learned feature distribution and uses this information to guide filtering. The resulting single-pass framework couples part segmentation with an explicit assessment of measurement reliability.

We first evaluate GUARD on a real-world HDD point cloud dataset collected from practical scanning and reconstruction pipelines, where sensor noise and reconstruction-induced ghost artifacts directly affect part-level perception for automated component handling and disassembly. This dataset serves as our primary engineering evaluation and reflects the types of geometric corruption encountered in realistic robotic deployment. To further assess robustness beyond the HDD setting, we conduct controlled experiments on ShapeNetPart~\cite{shapenet2015} with synthetic sensor noise and ghost-cluster corruption, and extend the evaluation to ScanNet~\cite{dai2017scannet} to examine the behavior of the proposed uncertainty modeling in large-scale indoor scenes. Across these settings, GUARD consistently improves robustness to corrupted geometry while maintaining competitive segmentation accuracy on cleaner inputs, demonstrating its potential as a practical perception component for robotic systems operating under imperfect 3D sensing.

Our contributions can be summarized as follows:
\begin{itemize}
\item \textbf{Reliability-aware interpretation of engineering point clouds.} We introduce a framework that couples part segmentation with unreliable-point filtering, addressing the distinction between recognizing a component and determining whether its observed geometry is trustworthy. The framework targets structured scanning and reconstruction artifacts encountered in real HDD point clouds.

\item \textbf{Geometry-conditioned uncertainty modeling.} We develop a multi-scale geometric representation and a multi-bandwidth random Fourier feature Gaussian Process that estimates uncertainty in geometric feature space. Fusion with predictive entropy combines evidence about geometric reliability and semantic ambiguity within a single-pass architecture that can be integrated with different segmentation backbones.

\item \textbf{Empirical analysis of the denoising--segmentation tradeoff.} Through evaluation on real HDD scans and complementary benchmarks, we show that geometric uncertainty and semantic confidence provide different information about corrupted measurements. Backbone comparisons and filtering analyses characterize when artifact removal improves segmentation and when removal of informative points can limit those gains, providing evidence for selecting reliability-aware filtering strategies.

\end{itemize}

\section{Related Work}
\noindent\textbf{Engineering information for robotic disassembly}
Recent research in engineering informatics has investigated robotic disassembly from the perspectives of human--robot collaboration, sequence planning, and manufacturing-system optimization.
For example, Guo et al.~\cite{guo2023human} studied human--robot collaborative disassembly sequence planning, Liang et al.~\cite{liang2023improved} addressed disassembly line balancing, and Zhang et al.~\cite{zhang2024mixed} investigated responsive collaborative disassembly optimization.
These studies primarily focus on planning and coordination within remanufacturing and disassembly systems.
Complementary to these efforts, the present study focuses on the reliability of the geometric information used to interpret components during robotic disassembly.

\noindent\textbf{Point Cloud Segmentation.}
Deep learning on point clouds has evolved rapidly since the introduction of PointNet++~\cite{qi2017pointnet++} and DGCNN~\cite{wang2019dgcnn}, which establish hierarchical and graph-based paradigms for local feature aggregation. 
Most subsequent advances, including attention-based frameworks and robustness-oriented methods, build upon these backbones. 
For example, PointNeXt~\cite{qian2022pointnext} revisits PointNet++ with improved training and scaling strategies, while attention-based variants such as UPA~\cite{xiu2022upa}, SOE-Net~\cite{li2021soenet}, and PointCloud-At~\cite{umar2024pointcloudat} introduce self-attention to enhance local feature modeling. 
Recent models such as Point Transformer~\cite{wu2024ptv3}, although attention-driven, still follow the hierarchical neighborhood aggregation paradigm introduced by PointNet++ and DGCNN, and can be viewed as an attention-based extension of these backbones.
As a result, PointNet++ and DGCNN remain representative foundations for modern point cloud segmentation networks and plug-in robust segmentation methods.
We therefore adopt them as benchmarks for fair comparison, and further evaluate our method on the most advanced Point Transformer backbone to verify its generalization.

\noindent\textbf{Robust and Noise-Aware Learning.}
Several approaches have been proposed to improve robustness under noisy or corrupted point clouds. 
PointCleanNet~\cite{rakotosaona2019pointcleannet} learns to restore clean shapes through supervised noise--clean correspondences, while PointCVaR~\cite{li2024pointcvar} leverages conditional value-at-risk optimization to emphasize hard or corrupted samples. 
REAL3D-AD~\cite{liu2023real3dad} and PO3AD~\cite{ye2025po3ad} further explore uncertainty-guided 3D anomaly detection by reconstructing clean geometry and identifying abnormal regions through residual-based discrepancies. 
Although conceptually related, their reconstruction-driven formulation targets anomaly localization rather than general noise suppression, and struggles to handle fine-grained geometric perturbations or ghost-cluster misalignment. 
Overall, these methods either require explicit noise annotations or involve multiple forward passes and resampling, making them computationally expensive and less suitable for real-time applications.

\noindent\textbf{Uncertainty Estimation.}
Uncertainty modeling provides a principled way to handle distributional shifts and ambiguous inputs. 
Bayesian deep learning~\cite{gal2016dropout, kendall2017uncertainties} and ensemble methods~\cite{lakshminarayanan2017simple} offer reliable uncertainty estimates but are computationally expensive. 
Spectral-normalized neural Gaussian processes (SNGP)~\cite{liu2020simple} introduce scalable uncertainty estimation by combining random Fourier features with spectral normalization, enabling efficient, single-pass predictive variance computation. 
Recent works integrate similar uncertainty modules into 2D vision tasks~\cite{van2020uncertainty}, but their application to large-scale 3D segmentation remains underexplored.

\noindent\textbf{Summary.}
Prior research has established complementary advances in engineering informatics for robotic disassembly, point-cloud segmentation and denoising, and scalable uncertainty estimation. 
The remaining challenge addressed in this work is how to integrate these capabilities when corrupted measurements form locally plausible structures that may still receive confident semantic predictions. 
GUARD addresses this issue by modeling uncertainty in learned geometric representations and combining it with semantic entropy to guide reliability-aware filtering during segmentation. 
This formulation provides a reliability-aware representation of imperfect geometric information, supporting more robust part-level interpretation of engineering point clouds affected by structured reconstruction artifacts.

\section{Methodology}
\subsection{Overview}

\begin{figure*}[!t]
    \centering
    \includegraphics[width=0.95\linewidth]{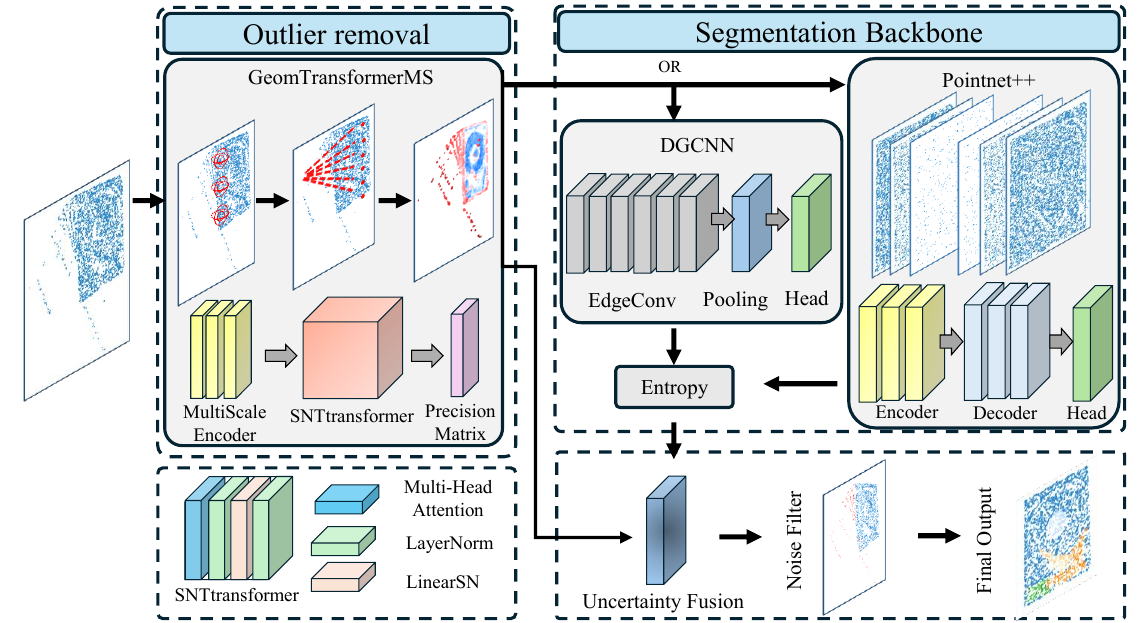}
    \caption{
    Overall architecture of the proposed framework. 
    Our method introduces an \textbf{outlier-removal head} composed of a multi-scale geometric encoder, 
    a geometry-aware transformer block, 
    and an uncertainty-aware Gaussian Process (GP) head. 
    As illustrated, it can be seamlessly attached to various backbones (e.g., PointNet++, DGCNN), 
    providing a unified plug-in module for robust and uncertainty-aware point cloud segmentation.
    }
    
    \label{fig:network}
\end{figure*}

Our goal is to achieve robust and uncertainty-aware point cloud segmentation under noisy perturbations and structured reconstruction artifacts. 
To this end, we propose a hybrid architecture that integrates a multi-scale geometric encoder with a Gaussian process--based uncertainty head designed for geometric feature-space modeling. 
Unlike prior SNGP implementations that attach the Gaussian process module to classifier logits, our module operates on intermediate geometric embeddings, allowing it to assign higher uncertainty to geometrically inconsistent structures such as ghost clusters or misaligned fragments. 
However, geometric uncertainty alone may not fully capture fine-grained ambiguity near object boundaries. 
Therefore, we further introduce a lightweight logit-entropy measure at the final prediction layer to characterize residual semantic uncertainty. 
This dual-uncertainty design enables the model to jointly reason about geometric reliability and semantic ambiguity, thereby improving robustness across different corruption conditions. The overall framework is illustrated in Fig.~\ref{fig:network}.

Within the proposed framework, relative offsets and distances characterize local geometric relationships, while the geometry-aware transformer captures broader spatial context in the learned feature representation. 
The Gaussian process--based uncertainty head estimates the reliability of these geometric representations, whereas predictive entropy reflects ambiguity in the corresponding semantic predictions. 
Their combination provides a point-wise reliability measure that is used to filter unreliable measurements, while the segmentation backbone produces the associated part labels. 
In this way, GUARD links geometric representation, reliability assessment, and semantic interpretation, transforming imperfect point-cloud measurements into more reliable part-level information for downstream robotic disassembly. 
This information-quality perspective is particularly important for engineering scans in which structured artifacts may appear geometrically plausible and can otherwise be retained as valid components.

\subsection{Geometric Feature Extraction}
This stage extracts both local and global geometric representations that are 
robust to noise while preserving fine structural details. 
Because the network operates directly on raw coordinates $(x,y,z)$, 
the input feature space is highly unstable---small positional jitters can cause large activation shifts in the first layer and even lead to training divergence. 
To stabilize this process, we apply spectral normalization (SN) to the layer's weight, 
which explicitly constrains its spectral norm and bounds the layer's Lipschitz constant:
\begin{equation}
    W_{\text{SN}} = \frac{W}{\sigma(W)}, 
    \qquad \sigma(W)=\max_{\|v\|_2=1}\|Wv\|_2.
\end{equation}
This normalization ensures that the mapping from coordinates to latent features 
has a controlled sensitivity to input perturbations. 
Unlike prior multi-scale or transformer-based encoders~\cite{qi2017pointnet++,wang2019dgcnn}, 
our design applies SN directly to the raw-coordinate projection, providing a theoretically grounded stabilization that improves uncertainty calibration in the downstream noise detection head.

\noindent\textbf{Local geometric features.}
For each point $x_i$, we build multi-scale neighborhoods 
$\mathcal{N}_{k_s}(i)$ with $k_s\!\in\!\{16,32,64\}$ to capture both fine and coarse geometry. 
To manage the computational overhead, we employ a chunked-KNN strategy 
that divides the neighbor search into smaller batches without altering topology, 
reducing GPU memory before the transformer stage. 
Each neighbor $x_j\!\in\!\mathcal{N}_{k_s}(i)$ is described by relative offset and distance:
\begin{equation}
\scalebox{0.9}{$
\boldsymbol{\delta}_{ij}=\mathbf{x}_j-\mathbf{x}_i,\quad
d_{ij}=\|\boldsymbol{\delta}_{ij}\|_2,\quad
\mathbf{f}_{ij}^{(s)}=[\,\boldsymbol{\delta}_{ij};\,d_{ij}\,]\in\mathbb{R}^4.
$}
\end{equation}

Each scale uses an independent MLP $\mathrm{MLP}^{(s)}:\mathbb{R}^4\!\rightarrow\!\mathbb{R}^{W}$ 
to embed $\mathbf{f}_{ij}^{(s)}$, followed by mean and max pooling over neighbors and concatenation across scales:

\begin{equation}
\mathbf{u}_i=\big[\bar{\mathbf{g}}_{i}^{(1)},\tilde{\mathbf{g}}_{i}^{(1)},\dots,
                 \bar{\mathbf{g}}_{i}^{(S)},\tilde{\mathbf{g}}_{i}^{(S)}\big].
\end{equation}

A spectrally normalized linear projection maps $\mathbf{u}_i$ into the embedding space:
\begin{equation}
\mathbf{h}_i = \mathrm{LN}\!\Big(\mathrm{Drop}\big(W_{\text{SN}}\mathbf{u}_i+b\big)\Big),
\end{equation}
where $W_{\text{SN}}$ is the normalized weight defined above, LN denotes LayerNorm, and Drop refers to dropout. 
This first-layer normalization stabilizes gradient flow and prevents large activation magnitudes 
caused by coordinate noise, enabling more consistent feature learning.

\noindent\textbf{Global geometric features.}
While local encoding preserves fine structural details, ghost clusters can exhibit locally plausible geometry yet contradict global spatial consistency.
To capture holistic context, we adopt a geometry-aware transformer that models interactions among all points through multi-head self-attention:
\begin{equation}
\mathrm{Attn}(\mathbf{Q},\mathbf{K},\mathbf{V})
= \mathrm{Softmax}\left(\frac{\mathbf{Q}\mathbf{K}^\top}{\sqrt{d}}+\Phi(\Delta)\right)\mathbf{V},
\end{equation}
where $\Phi(\Delta)$ encodes pairwise geometric relations, such as relative offsets and distances, injecting spatial awareness directly into the attention weights.
We stack $L$ transformer blocks with pre-norm residual connections, allowing the network to propagate long-range dependencies while maintaining numerical stability.
To further stabilize training on raw 3D coordinates, we apply spectral normalization to the first transformer block, ensuring consistent gradient scaling across modules.
A final LayerNorm produces globally contextualized representations
$\mathbf{H}^{(L)}=\mathrm{LN}\big(\mathbf{H}^{(L)}\big)$ that fuse local precision with global coherence.
These stabilized geometric embeddings are then fed into the Gaussian Process--based uncertainty head for subsequent segmentation.

\subsection{Multi-bandwidth Geometric Gaussian Process Head (GGPH)}
\label{sec:geo-gp-head}
\begin{figure*}[!t]
    \centering
    \subfloat[Conceptual illustration of the geometric uncertainty hypothesis.]{%
        \includegraphics[height=3.9cm]{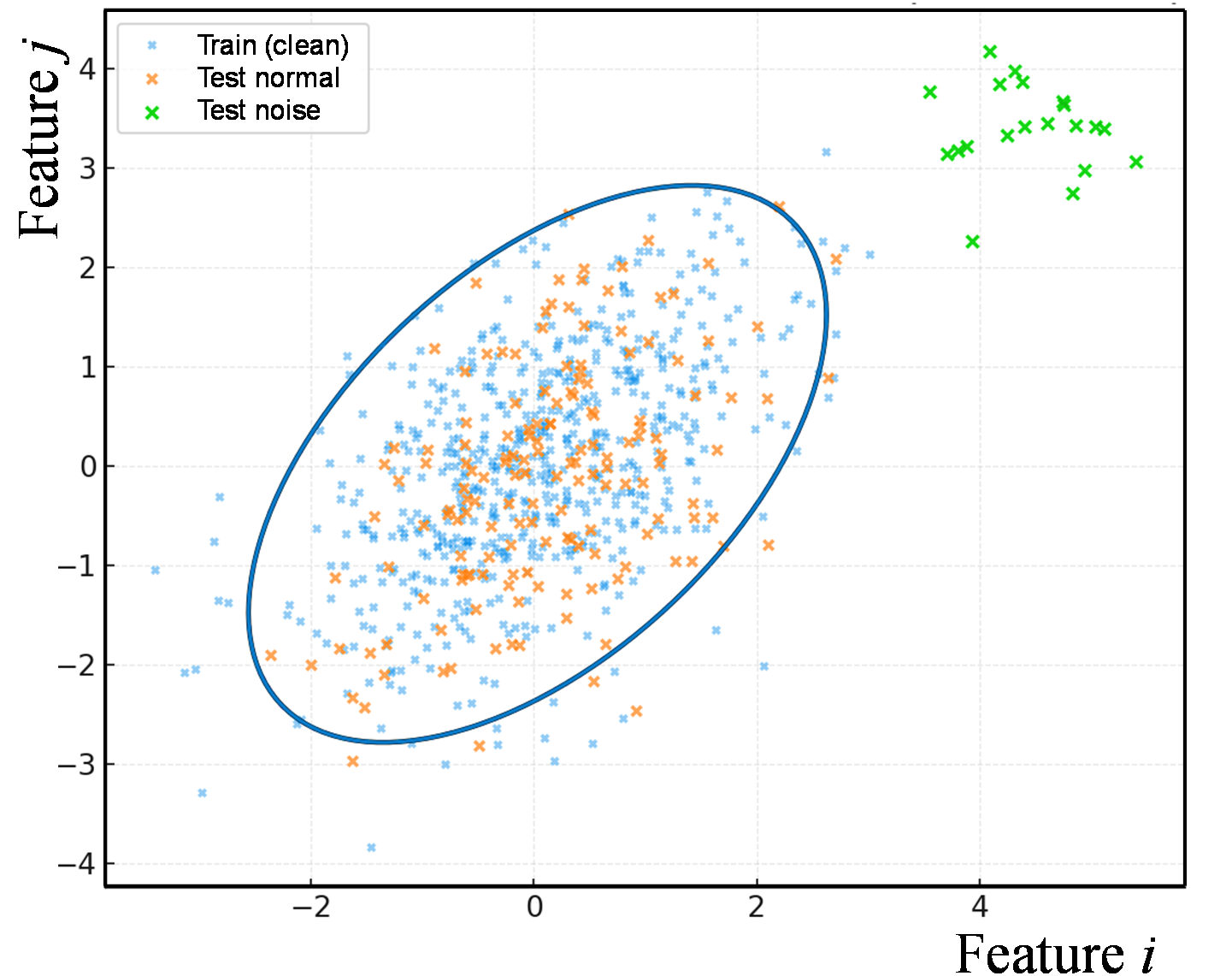}}
    \hspace{0.6cm}
    \subfloat[Empirical variance distribution on the HDD dataset.]{%
        \includegraphics[height=4.2cm]{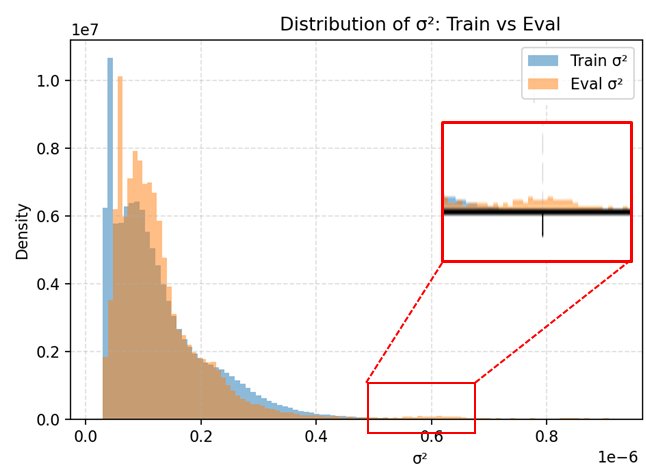}}
    \caption{\textbf{Motivation for geometric uncertainty modeling.}
    (a) We hypothesize that clean points lie near a compact geometric manifold in the embedding space, while outlier or ghost-cluster points deviate from this manifold, leading to higher predictive uncertainty.
    (b) The predicted variance $\sigma^2$ distribution on the HDD dataset supports this hypothesis: most samples concentrate in a low-uncertainty region, while deviations correspond to abnormal or noisy structures.}
    \label{fig:hypo_test}
\end{figure*}

While the geometric transformer captures global structure, 
it remains deterministic and lacks the ability to assess confidence 
when encountering unseen or noisy geometry.  
To illustrate this challenge, we visualize the learned geometric feature space in Fig.~\ref{fig:hypo_test}.

We address this limitation by modeling geometric uncertainty through a Multi-bandwidth Random Fourier Feature Gaussian Process (MultiRFF-GP) head, which extends SNGP~\cite{liu2020simple} to point-cloud geometry. 
While the preceding geometric transformer captures global context deterministically, 
it lacks the ability to quantify epistemic uncertainty caused by unseen or noisy structures such as ghost clusters. 
The MultiRFF-GP head fills this gap by approximating a Gaussian process kernel in the geometric feature space 
using random Fourier features sampled at multiple frequency bands. 
This multi-bandwidth design enables the model to express both short-range geometric variation 
and long-range structural uncertainty within a unified probabilistic framework.

Given the geometric feature $\mathbf{h}_i\!\in\!\mathbb{R}^{C_t}$, 
we first construct random Fourier features with three characteristic bandwidths 
$\sigma_l\!\in\!\{0.3,0.7,1.4\}$, each capturing a distinct spatial scale:
\begin{equation}
\scalebox{0.9}{$
\phi_i = 
\sqrt{\tfrac{2}{M}}
\cos\!\left(\mathbf{h}_i W + b\right),
\quad
W = [W^{(1)},W^{(2)},W^{(3)}],
$}
\end{equation}

where $W^{(l)}\!\sim\!\mathcal{N}(0,\sigma_l^{-2}I)$ and $b^{(l)}\!\sim\!\mathcal{U}(0,2\pi)$.
Concatenating the features across bandwidths yields a rich spectral embedding 
that jointly represents local shape detail and global geometry.

During training, we maintain an online estimate of the feature second-moment matrix used to construct the GP precision matrix:
\begin{equation}
P \leftarrow P + \Phi^\top \Phi,
\end{equation}
where $\Phi$ stacks the RFF embeddings in each batch. At inference, we form the regularized precision matrix and compute the predictive variance as
\begin{align}
K &= P+\lambda I = LL^\top,\
\end{align}
\begin{align}
u_i &= \mathrm{CholeskySolve}(L,\phi_i^\top)
\end{align}
\begin{align}
\sigma_i^2=\phi_i u_i
=\phi_iK^{-1}\phi_i^\top.
\end{align}
The variance $\sigma_i^2$ reflects geometric uncertainty, with high values indicating regions poorly supported by the training distribution, such as outliers or ghost clusters.

Unlike prior GP-based uncertainty heads, our approach incorporates 
\emph{multiple kernel bandwidths}, enabling fine-to-coarse uncertainty modeling 
across geometric scales. 
Moreover, by operating directly on the spectrally normalized geometric embeddings from the encoder, 
the MultiRFF-GP head inherits stable feature scaling, which improves the conditioning of the precision matrix 
and leads to more reliable uncertainty calibration. 
Together, these properties make the MultiRFF-GP head a key component of our framework, effectively bridging deterministic geometric reasoning with probabilistic uncertainty estimation 
for robust point cloud segmentation.

\subsection{Hybrid Uncertainty Fusion for Robust Segmentation}
\label{sec:hybrid-uncertainty}

The variance estimated by the MultiRFF-GP head effectively identifies 
major outliers that lie far from the learned geometric manifold. 
These large \textit{ghost clusters}, often caused by repeated scans or global misalignments, 
exhibit clearly inconsistent structures and are well separated from valid parts in the feature space. 
However, not all ghost clusters are distant. 
Some minor clusters appear very close to real object surfaces---spatially overlapping with valid geometry and sharing similar local patterns. 
Even for humans, such near-surface outliers are difficult to judge: 
they may or may not be manually removed across datasets, resulting in inconsistent data quality. 
Consequently, the GP head, which focuses on large-scale geometric deviations, 
becomes less sensitive to these small but semantically confusing artifacts.

In contrast, the predictive entropy of the softmax logits reflects the model's 
uncertainty about semantic consistency. 
When a point lies close to the true surface but carries mixed or duplicated local features, 
its neighborhood generates fluctuating logits across multiple parts, producing higher entropy than normal points. 
Therefore, entropy naturally captures these boundary-level or label-like ambiguities that variance alone cannot highlight.

\begin{figure*}[!t]
    \centering
    \subfloat[HDD scan dataset\label{fig:noise_hdd}]{%
        \includegraphics[width=0.42\textwidth, trim=15 10 15 10, clip]{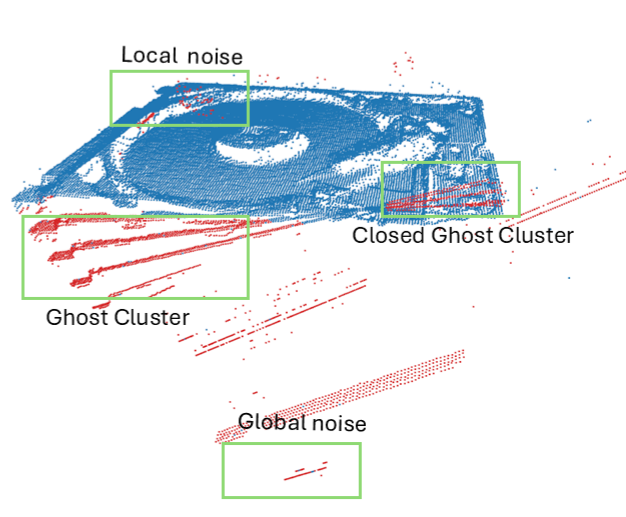}}
    \hspace{0.6cm}
    \subfloat[ScanNet dataset\label{fig:noise_scannet}]{%
        \includegraphics[width=0.42\textwidth]{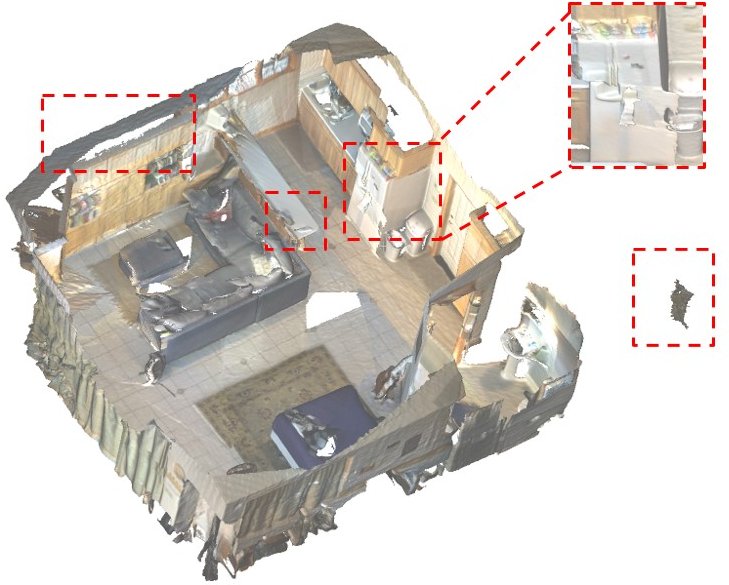}}
    \caption{Visualization of representative noise.}
    \label{fig:noise_visualization}
\end{figure*}

As illustrated in Fig.~\ref{fig:noise_visualization}, 
the ghost clusters manifest as globally inconsistent duplicate structures, 
whereas closed ghost cluster corresponds to locally confusing regions near the part surface. 
To account for both types of ghost clusters, we combine the normalized GGPH variance $\tilde{\sigma}_i^2$ and the normalized predictive entropy $\tilde{H}_i$ of the softmax logits into a unified hybrid score. The \textit{far ghost clusters}, which appear as detached or globally inconsistent structures, lie outside the training manifold and are effectively detected by the GGPH predictive variance. On the other hand, the \textit{close ghost clusters}, which reside near valid surfaces and exhibit overlapping local geometry, induce ambiguous logits and are more sensitively captured by the softmax entropy. Since both forms of ghost clusters coexist and exhibit correlated yet distinct uncertainty characteristics, we adopt weighted combination coefficients rather than a cascade scheme, providing a smooth trade-off between epistemic and aleatoric components. The final hybrid uncertainty is defined as
\begin{equation}
U_i = \alpha\,\tilde{\sigma}_i^2 + (1 - \alpha)\,\tilde{H}_i,
\label{eq:hybrid_uncertainty}
\end{equation}
where $\alpha \in [0,1]$ balances the contribution of geometric out-of-distribution uncertainty and semantic ambiguity.

We set $\alpha{=}0.7$ by default, emphasizing geometric uncertainty while 
retaining a lighter entropy contribution to filter ambiguous boundary noise.
This fusion forms a hybrid uncertainty filtering mechanism: the GGPH variance suppresses global ghost clusters, and the entropy term further cleans residual local artifacts.

The resulting hybrid uncertainty map provides a spatially coherent confidence field 
that aligns well with both structural and semantic noise patterns, 
allowing the model to downweight unreliable points 
and achieve robust segmentation even under severe scanning perturbations.

\subsection{Implementation Details}
We implement GUARD in PyTorch, integrating the MultiRFF-GP head into both PointNet++ and DGCNN backbones. 
During training, the precision matrix is accumulated online to capture the evolving feature distribution and is frozen during testing for efficient inference.
Further optimization and dataset-specific configurations are described in Section~\ref{sec:exp-settings}.

\section{Experiments}

The evaluation is organized around three main questions: 
(1) whether geometric uncertainty can distinguish corrupted measurements from valid component geometry; 
(2) whether uncertainty-guided filtering improves part segmentation on real HDD scans; and 
(3) how filtering affects the tradeoff among artifact removal, preservation of informative structures, and computational efficiency.
The HDD dataset provides the primary application-specific evaluation, while ShapeNetPart enables controlled analysis across different corruption types and severities, and ScanNet examines the behavior of the proposed uncertainty modeling in a different real-world reconstruction setting.
Together, these complementary experiments evaluate the reliability, robustness, and generality of the proposed perception framework across application-specific and benchmark settings.

\subsection{Experiment Settings}
\textbf{Dataset}   
We evaluate our method on three datasets covering real-world scans, standard object-level benchmarks, and large-scale indoor scenes. First, we construct a real-scan dataset of hard disk drives (HDDs), which contains 2,745 point clouds collected from different brands and models. The diversity in brand and design introduces substantial shape variations, making the dataset challenging. We randomly split the dataset into 80\% for training and 20\% for testing.

In addition, we adopt the widely used ShapeNetPart benchmark~\cite{shapenet2015} for part segmentation, which consists of 12,128 training samples and 2,874 testing samples across 16 object categories. This benchmark enables direct comparison with existing works in point cloud segmentation. Our primary benchmark focuses on part segmentation rather than large-scale semantic segmentation benchmarks such as S3DIS, since our target application emphasizes fine-grained object understanding for robotic manipulation tasks (e.g., desktop disassembly). In these scenarios, recognizing detailed object parts is more critical than scene-level semantics, making ShapeNetPart a more suitable benchmark. Together with our real-scan HDD dataset, which captures domain-specific challenges such as brand variations and scanning noise, ShapeNetPart provides a complementary evaluation: the former reflects real-world robotic use cases, while the latter offers a controlled and widely recognized benchmark for fair comparison.

To further validate robustness and generality, we evaluate our method on ScanNet~\cite{dai2017scannet}, a large-scale indoor scene dataset with complex layouts and reconstruction artifacts. 
This experiment examines whether our uncertainty-guided filtering remains effective under challenging real-world conditions.

\noindent\textbf{Noise Injection.}  
To evaluate robustness, we further inject synthetic noise into ShapeNetPart following the protocol of PointCloud-C~\cite{sun2022benchmarking}(local and global noise), which systematically introduces outliers under five severity levels (5\%, 10\%, 20\%, 30\%, and 40\%). These severity ratios are not arbitrarily chosen: our analysis of the HDD dataset shows that the proportion of noisy points in real scans typically ranges from about 5\% in the cleanest cases to nearly 40\% in the noisiest ones. Unlike the original setup that replaces a fraction of points with outliers while keeping the total count fixed, we found from real HDD scans that noise more commonly appears as \emph{extra points} rather than missing ones---since data acquisition often favors capturing more points to avoid losing information. Therefore, we revise the PointCloud-C procedure by appending additional noisy points instead of replacement, yielding a noise distribution closer to real-world scans. In our experiments, we particularly highlight the 5\% (least) and 40\% (most) cases, which directly reflect the realistic noise levels observed in practice.

Beyond random local or global perturbations, we further introduce a \emph{ghost-cluster noise} that replicates the duplicated structures frequently observed in real HDD scans. Each ghost cluster is generated by cloning a subset of points from a dominant object part and applying a small rigid transformation consisting of random rotation and translation. The severity level controls the number and spatial deviation of these clusters: higher levels produce more duplicated parts with larger displacements. This process yields globally inconsistent but locally coherent structures, closely matching the ``double-scanned'' artifacts commonly seen in practical industrial 3D sensing.

\noindent\textbf{Training and Evaluation.}\label{sec:exp-settings}
All models are implemented in PyTorch, trained on a single NVIDIA A100 GPU, and evaluated on an NVIDIA RTX 4080 for inference.
We adopt both PointNet++ and DGCNN backbones, upon which our proposed GUARD module is integrated. 
For PointNet++, we train for 200 epochs with a batch size of 16 using the Adam optimizer 
(initial learning rate $1\times10^{-3}$, decayed by 0.5 every 20 epochs). 
For DGCNN, we train for 200 epochs with a batch size of 32 using SGD with momentum 0.9, 
an initial learning rate of 0.1, and a cosine annealing schedule with a minimum learning rate of $1\times10^{-3}$. 
A weight decay of $1\times10^{-4}$ is applied in all experiments. 
The random feature dimension is set to $D=1024$, and the precision matrix is accumulated online during training and frozen for testing. 
Input point clouds are uniformly sampled to 2,048 points for ShapeNetPart and 2,500 points for HDD scans. 
We report instance mIoU, ratio of noise points retained, and inference latency for evaluation.

\subsection{Main Results}

\subsubsection{Noise sensitivity of backbones}  
We first evaluate two widely used segmentation backbones, PointNet++ and DGCNN, under both clean and noisy inputs. 
As shown in Table~\ref{tab:noise_baseline}, both models undergo substantial performance degradation when noise is present. For instance, under ghost-cluster noise, PointNet++ drops from 0.8449 mIoU (clean) to 0.6082, and DGCNN from 0.8501 to 0.6200.
This confirms that existing architectures are highly vulnerable to outliers, motivating the need for plug-in modules to enhance robustness without redesigning the backbone.

\begin{table}[!tb]
\centering
\scriptsize                
\setlength{\tabcolsep}{3pt} 
\renewcommand{\arraystretch}{1.0} 
\caption{Baseline comparison under different noise types.}
\label{tab:noise_baseline}
\begin{tabular}{lccccccc}
\toprule
Backbone & Clean & Ghost & Drop (\%) & Global & Drop (\%) & Local & Drop (\%) \\
\midrule
PointNet++ & 0.8449 & 0.6082 & $-28.0$ & 0.3242 & $-61.6$ & 0.5073 & $-40.0$ \\
DGCNN      & 0.8501 & 0.6200 & $-27.1$ & 0.4310 & $-49.3$ & 0.5582 & $-34.3$ \\
\bottomrule
\end{tabular}
\end{table}

\begin{table*}[!t]
\centering
\caption{Robustness under different noise types. HDD/ScanNet (real) mIoU is reported separately. The ShapeNet block reports mIoU at 5\% (S1) and 40\% (S5) severity for \textbf{GhostCluster}, \textbf{Global}, and \textbf{Local} noise, plus averages (\cmIoUGC, \cmIoUG, \cmIoUL, overall \cmIoUA).}
\label{tab:robust_miou_extended}

\resizebox{\linewidth}{!}{%
\begin{tabular}{ll|c|c|cc|cc|cc|cccc}
\toprule
\multirow{2}{*}{Backbone} & \multirow{2}{*}{Method} &
\multirow{2}{*}{HDD mIoU$\uparrow$} &
\multirow{2}{*}{ScanNet mIoU$\uparrow$} &
\multicolumn{10}{c}{\textbf{ShapeNetPart (synthetic corruptions; S1=5\%, S5=40\%)}} \\
\cmidrule(lr){5-14}
 &  &  &  &
\multicolumn{2}{c|}{GhostCluster mIoU$\uparrow$} &
\multicolumn{2}{c|}{Global mIoU$\uparrow$} &
\multicolumn{2}{c|}{Local mIoU$\uparrow$} &
\multicolumn{4}{c}{\textbf{Averages over ShapeNet only} $\uparrow$} \\
 &  &  &  & S1 & S5 & S1 & S5 & S1 & S5 & \cmIoUGC & \cmIoUG & \cmIoUL & \cmIoUA \\
\midrule
PointNet++ & Vanilla         & 0.7739 & 0.5259 & 0.8043 & 0.6082 & 0.6543 & 0.3242 & 0.8091 & 0.5073 & 0.7175 & 0.4729 & 0.6757 & 0.6221 \\
           & +PointCVaR      & 0.6685 & 0.5281 & 0.7840 & 0.6049 & 0.7244 & 0.4125 & 0.7851 & 0.5207 & 0.7049 & 0.5699 & 0.6672 & 0.6473 \\
           & +\textbf{GUARD} & \textbf{0.8318} & \textbf{0.5601} & \textbf{0.8176} & \textbf{0.7014} & \textbf{0.8111} & \textbf{0.4409} & \textbf{0.8206} & \textbf{0.6131} & \textbf{0.7670} & \textbf{0.6396} & \textbf{0.7220} & \textbf{0.7095} \\
\midrule
DGCNN      & Vanilla         & \textbf{0.7434} & 0.4865 & 0.7878 & 0.6200 & 0.7421 & 0.4310 & 0.7902 & 0.5582 & 0.7139 & 0.5879 & 0.6805 & 0.6608 \\
           & +PointCVaR      & 0.6964 & 0.4867 & 0.8001 & 0.6555 & 0.7106 & 0.4051 & 0.8032 & 0.6266 & 0.7378 & 0.5605 & 0.7205 & 0.6729 \\
           & +\textbf{GUARD} & 0.7425 & \textbf{0.5125} & \textbf{0.8293} & \textbf{0.7320} & \textbf{0.7471} & \textbf{0.5202} & \textbf{0.8258} & \textbf{0.6819} & \textbf{0.7855} & \textbf{0.6479} & \textbf{0.7530} & \textbf{0.7288} \\
\bottomrule
\end{tabular}%
}
\end{table*}

\subsubsection{Comparison with plug-in modules}  
We compare our approach with PointCVaR~\cite{li2024pointcvar}, the most recent plug-in module for outlier and noise removal that can be inserted into PointNet++ and DGCNN.
We evaluate on both the HDD scan data and ShapeNetPart with synthetic corruptions.  
As shown in Table~\ref{tab:robust_miou_extended}, GUARD achieves stronger overall robustness than the vanilla backbones and their PointCVaR-augmented counterparts.
On HDD, it improves mIoU by \textbf{7.5\%} over the vanilla PointNet++, highlighting strong generalization to real-world scans. 
On ShapeNetPart, our approach achieves up to \textbf{6--8\%} higher average mIoU under GhostCluster and Local noise, maintaining robustness even at 40\% corruption severity. 
These results confirm that our method not only enhances segmentation accuracy but also produces more stable and trustworthy predictions, which are critical for robotic applications.
For brevity, we report the overall results here, while the per-category performance is analyzed in the later experiment subsection.

\begin{table}[!tb]
\centering
\caption{Average noise kept (\%) under different corruption types.}
\label{tab:noise_kept}
\setlength{\tabcolsep}{4pt} 
\begin{tabular}{l l c c c c c}
\toprule
Type & Method & s1$\downarrow$ & s2$\downarrow$ & s3$\downarrow$ & s4$\downarrow$ & s5$\downarrow$ \\
\midrule
Local        & PointCVaR & 4.9 & 9.0 & 17.2 & 23.9 & 30.0 \\
             & GUARD     & 4.4 & 8.1 & 14.6 & 20.2 & 23.7 \\
\midrule
Global       & PointCVaR & 2.4 & 4.6 & 8.4  & 11.0 & 12.6 \\
             & GUARD     & 1.5 & 2.9 & 6.4  & 11.9 & 19.8 \\
\midrule
Ghostcluster & PointCVaR & 5.7 & 8.0 & 9.9  & 22.8 & 26.9 \\
             & GUARD     & 5.5 & 7.3 & 8.9  & 20.1 & 22.3 \\
\bottomrule
\end{tabular}
\end{table}
\subsubsection{Reliability and Efficiency}
In addition to robustness, practical deployment in robotic systems demands reliability and efficiency. 
As shown in Table~\ref{tab:noise_kept}, GUARD retains a smaller percentage of noisy points across most corruption types and severity levels. 
The improvement is most pronounced for \emph{GhostCluster} noise, reducing residual noise from 26.9\% to 22.3\% at S5, while also achieving steady gains under Local corruptions. These results indicate that our uncertainty-guided filtering effectively suppresses spurious points while preserving structural integrity, providing a strong foundation for downstream manipulation and segmentation tasks.

\begin{figure*}[!t]
  \centering
  \subfloat[Average inference latency comparison on an RTX~4080 GPU.\label{fig:latency_time}]{%
    \includegraphics[width=0.52\linewidth]{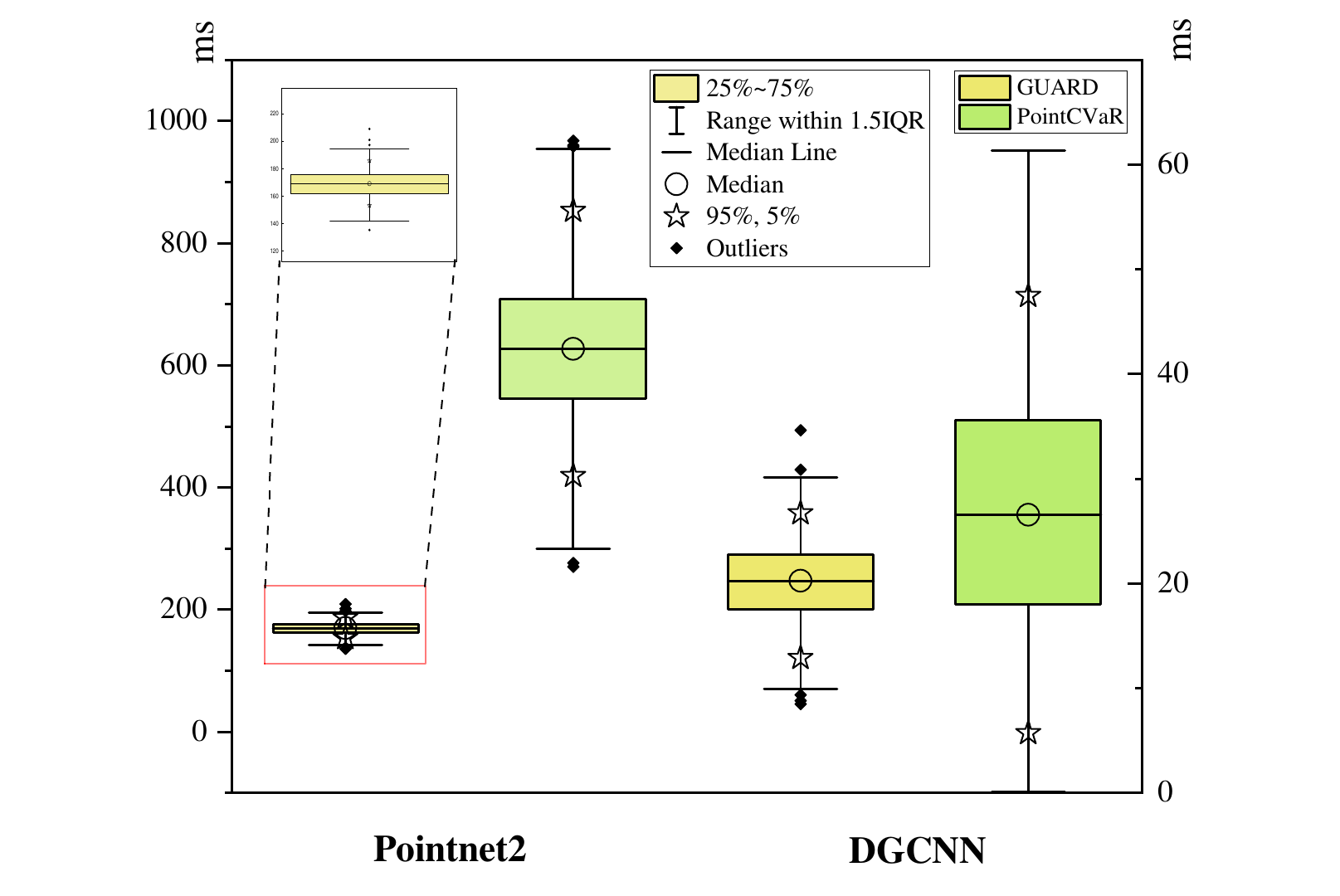}}
  \hfill
  \subfloat[Robustness comparison across different noise severities and types.\label{fig:noise_level}]{%
    \includegraphics[width=0.44\linewidth]{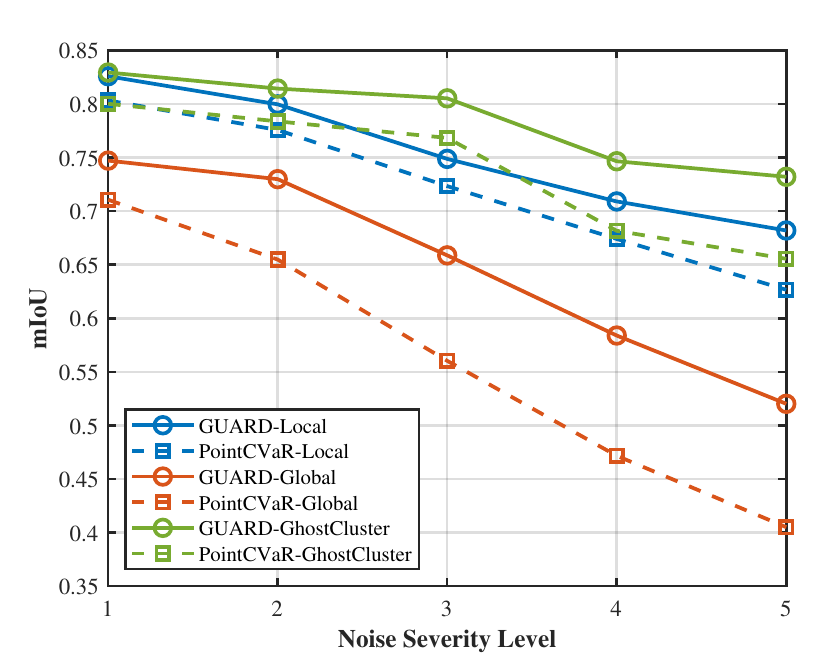}}

  \caption{\textbf{Efficiency and robustness evaluation of the proposed method.}
  (a) Inference latency comparison across different methods on an RTX~4080 GPU.
  (b) Segmentation robustness under varying noise types and severity levels.}
  
  \label{fig:latency_robustness}
\end{figure*}

Meanwhile, we also compare inference speed. As illustrated in Fig.~\ref{fig:latency_time}, our approach achieves efficient single-pass inference, with an average latency of ~169 ms per sample on PointNet++ and ~20 ms on DGCNN when tested on an RTX 4080 GPU. This is significantly faster than PointCVaR, which requires two forward passes (average 623 ms on PointNet++ and 26 ms on DGCNN) and exhibits higher tail latency (95th percentile 43.5 ms vs. ours 27.7 ms).
These results highlight that our single-pass design achieves a favorable balance between robustness and computational efficiency, making it well-suited for real-time robotic applications.

As shown in Fig.~\ref{fig:noise_level}, our method consistently surpasses PointCVaR across all noise types and severity levels.
For both local and global random noise, it maintains notably higher mIoU under increasing corruption,
showing strong resilience to spatial distortions and large-scale perturbations.
More importantly, our approach excels under ghost-cluster noise, where duplicated or ambiguous structures often cause severe misclassification.
While PointCVaR rapidly degrades, our uncertainty-guided filtering effectively distinguishes genuine surface points from mirrored clusters.

\subsubsection{Generalization to Point Transformer}

To further validate the generality of our method, we evaluate GUARD on the advanced backbone Point Transformer~(PTv3). 
Since PTv3 is a relatively heavy model that typically benefits from larger training datasets, we conduct experiments on the ScanNetV2 dataset, which is substantially larger and more diverse than ShapeNetPart and our HDD dataset. 
This setting allows the backbone to be trained more effectively and provides a realistic benchmark for evaluating robustness in large-scale indoor scenes. 
The quantitative results are summarized in Table~\ref{tab:cat_miou_scannet}.

\begin{table*}[!t]
\centering
\caption{Per-category mIoU results on ScanNet for PointTransformerV3 (PTv3) and our GUARD. Improved results are in bold.}
\label{tab:cat_miou_scannet}

\begin{tabular}{l|c|cccccccccc}
\toprule
\textbf{Method} & \textbf{mIoU} &
Wall & Floor & Cab. & Bed & Chair & Sofa & Table & Door & Window & Shelf \\
\midrule
PTv3
& 0.7757
& 0.8724 & 0.9568 & 0.7129 & 0.8343 & 0.9294 & 0.8623 & 0.7912 & 0.7499 & 0.7390 & 0.8302 \\

+\textbf{GUARD}
& \textbf{0.7902}
& \textbf{0.8785} & \textbf{0.9609} & \textbf{0.7208} & \textbf{0.8384} & \textbf{0.9372} & \textbf{0.8721} & \textbf{0.8043} & \textbf{0.7520} & \textbf{0.7445} & \textbf{0.8364} \\

\midrule
\textbf{Method} & \textbf{mIoU} &
Picture & Counter & Desk & Curtain & Fridge & Shower & Toilet & Sink & Bathtub & OtherFurn. \\
\midrule
PTv3
& 0.7757
& 0.3979 & 0.7168 & 0.7403 & 0.7896 & \textbf{0.7119} & \textbf{0.7244} & 0.9336 & \textbf{0.7095} & 0.8821 & 0.6292 \\

+\textbf{GUARD}
& \textbf{0.7902}
& \textbf{0.4018} & \textbf{0.7325} & \textbf{0.7421} & \textbf{0.7996} & 0.7043 & 0.7227 & \textbf{0.9614} & 0.6923 & \textbf{0.8896} & \textbf{0.6338} \\

\bottomrule
\end{tabular}
\end{table*}

\subsubsection{Uncertainty Quality}

\begin{figure*}[!t]
    \centering
    \includegraphics[width=0.45\linewidth]{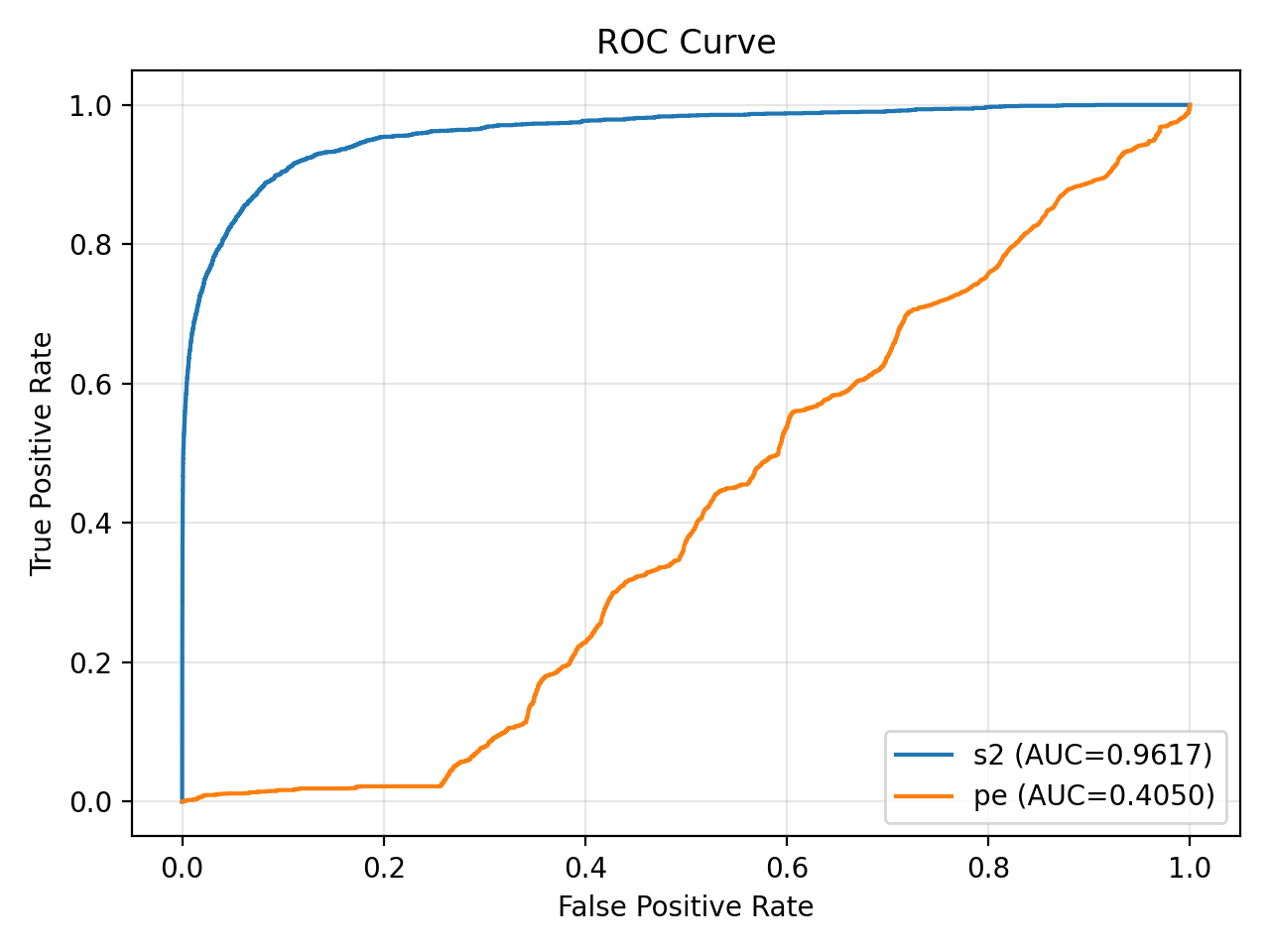}
    \hspace{0.6cm}
    \includegraphics[width=0.45\linewidth]{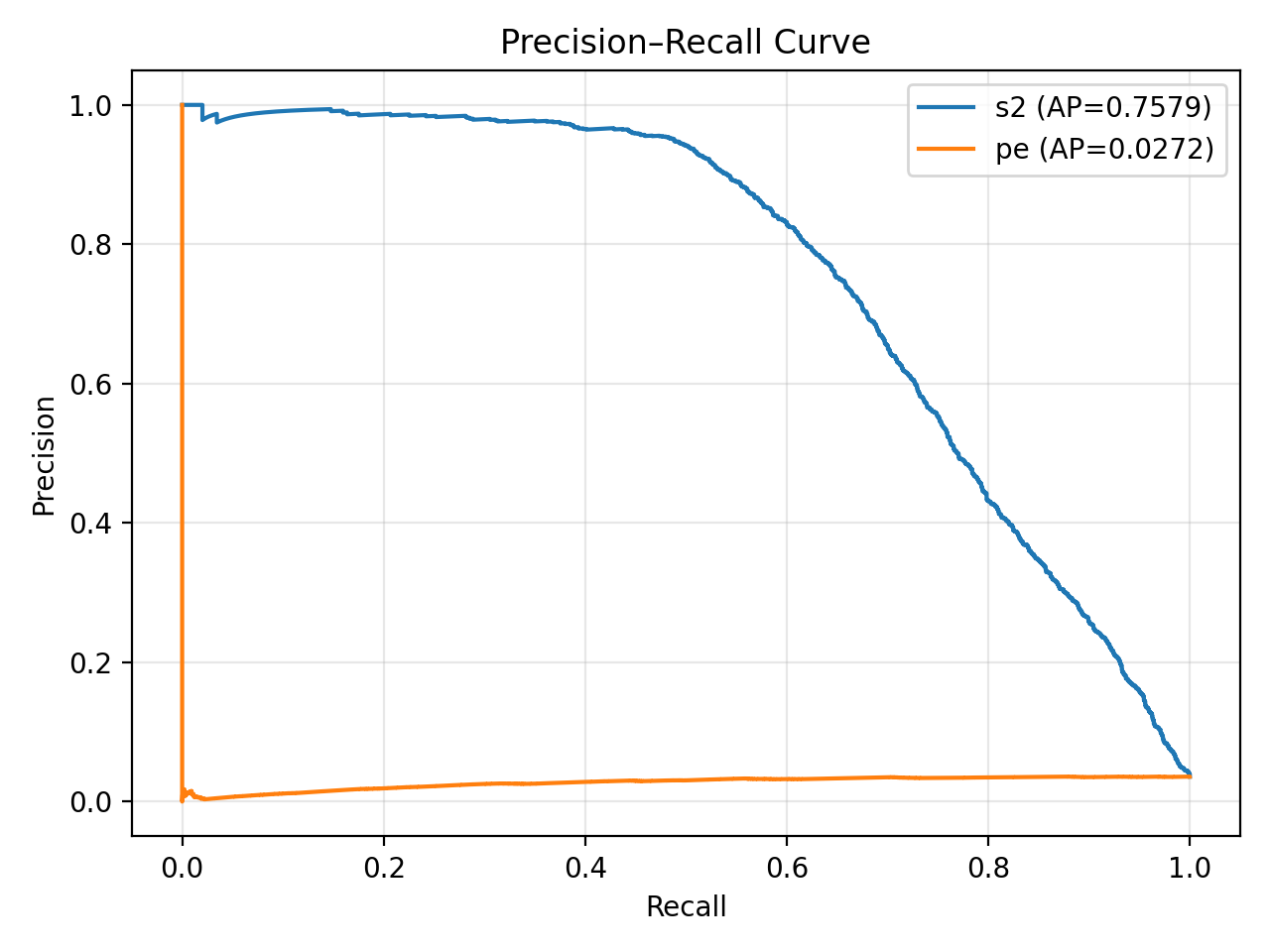}
    \caption{
        PR and ROC curves for noise detection.
    }
    \label{fig:roc}
\end{figure*}

To assess whether the estimated uncertainty can reliably identify corrupted points, Fig.~\ref{fig:roc} presents  ROC curves and PR curves for noise detection using two uncertainty measures: predictive entropy (PE) and the proposed geometry-conditioned uncertainty $\sigma_{\text{geo}}$ (S2, GUARD). Compared with PE, $\sigma_{\text{geo}}$ consistently achieves clearer separation between noisy and clean points, leading to significantly higher precision under class imbalance. This demonstrates that GUARD captures geometric unreliability more effectively than entropy-based uncertainty, which is mainly derived from semantic prediction ambiguity.

Since ScanNet does not provide explicit noise annotations, we manually label a subset of scenes to quantify denoising performance in large-scale indoor environments. For evaluation, we apply a filtering ratio of 0.9 and report noise detection performance in terms of Precision, Recall, and F1 score, as summarized in Table~\ref{tab:noise_detection}.

\begin{table}[!tb]
\centering
\caption{Noise detection performance on manually annotated ScanNet samples.}
\label{tab:noise_detection}
\begin{tabular}{lccc}
\toprule
Method & Precision $\uparrow$ & Recall $\uparrow$ & F1 Score $\uparrow$ \\
\midrule
Predictive Entropy (PE) & 0.2284 & 0.2144 & 0.2212 \\
GUARD ($\sigma_{\text{geo}}$) & 0.8032 & 0.7832 & 0.7931 \\
\bottomrule
\end{tabular}
\end{table}

\subsubsection{Model Parameter Analysis}

As reported in Table~\ref{tab:param_size}, GUARD adds only a lightweight parameter overhead of approximately 0.4M parameters to both DGCNN and PTv3. Despite this modest increase in model size, GUARD consistently improves both denoising quality and downstream segmentation performance. These results suggest that the observed gains mainly come from the proposed uncertainty-guided modeling strategy rather than from a substantial increase in model capacity.

\begin{table}[!tb]
\centering
\caption{Parameter size (M) comparison.}
\label{tab:param_size}
\small
\setlength{\tabcolsep}{3pt}
\resizebox{\columnwidth}{!}{%
\begin{tabular}{lccc}
\toprule
\textbf{Model}
& \textbf{DGCNN}
& \textbf{PointNet++}
& \textbf{Point Transformer} \tabularnewline
\midrule
\textbf{\# Params (M)}
& 1.9(+0.4)
& 2.2(+0.3)
& 46.2(+0.4) \tabularnewline
\bottomrule
\end{tabular}
}
\end{table}

\subsection{Visualization Results}

\begin{figure*}[!t]
    \centering
    \setlength{\tabcolsep}{3pt}
    \renewcommand{\arraystretch}{1.0}

    \newcommand{\hddimg}[1]{%
        \includegraphics[width=0.198\textwidth]{#1}%
    }

    \begin{tabular}{@{}c@{\hspace{5pt}}cccc@{}}
        \rotatebox{90}{\qquad \small\textbf{Original}} &
        \hddimg{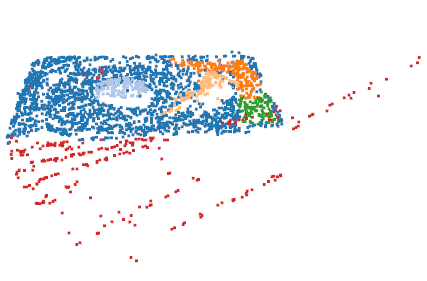} &
        \hddimg{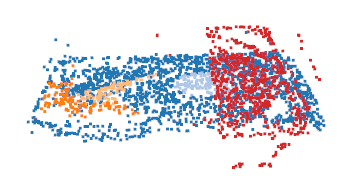} &
        \hddimg{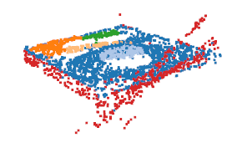} &
        \hddimg{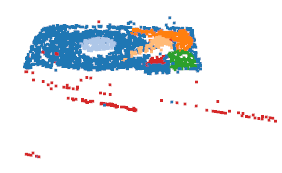} \\[5pt]

        \rotatebox{90}{\qquad \small\textbf{PointCVaR}} &
        \hddimg{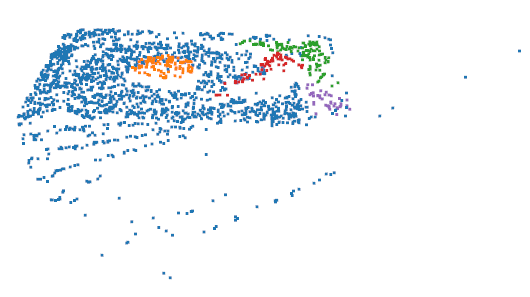} &
        \hddimg{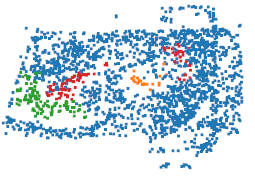} &
        \hddimg{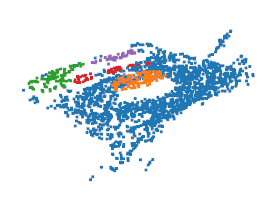} &
        \hddimg{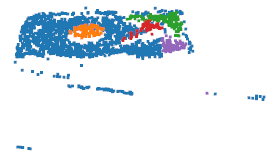} \\[5pt]

        \rotatebox{90}{\qquad \small\textbf{GUARD}} &
        \hddimg{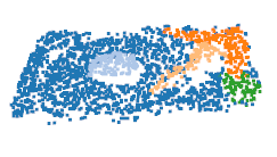} &
        \hddimg{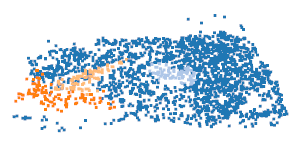} &
        \hddimg{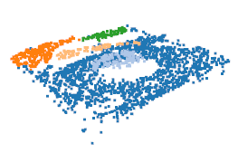} &
        \hddimg{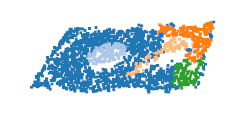}
    \end{tabular}

    \caption{Visualization of segmentation results on real HDD point clouds.}
    \label{fig:vis_hdd}
\end{figure*}

To qualitatively assess the effectiveness of our uncertainty-guided filtering,
we visualize segmentation results on both real and synthetic datasets.
Figure~\ref{fig:vis_hdd} presents results on real HDD scans, where our method effectively removes misaligned or duplicated clusters and restores fine geometric details,
demonstrating robust performance on held-out real HDD scans containing realistic scanning and reconstruction artifacts.

\begin{figure*}[!t]
    \centering
    \setlength{\tabcolsep}{2pt}
    \renewcommand{\arraystretch}{1.0}

    \resizebox{\linewidth}{!}{%
    \begin{tabular}{@{}p{10mm}cccccccc@{}}
        \parbox[t]{10mm}{\centering\rotatebox{90}{\scriptsize\textbf{  Original}}} &
        \includegraphics[height=14mm]{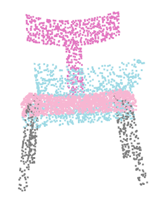} &
        \includegraphics[height=14mm]{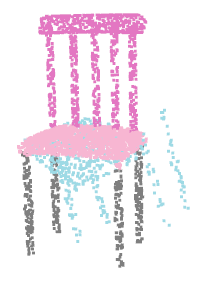} &
        \includegraphics[height=14mm]{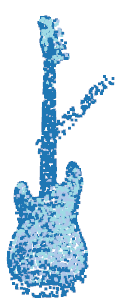} &
        \includegraphics[height=14mm]{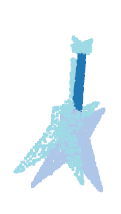} &
        \includegraphics[height=14mm]{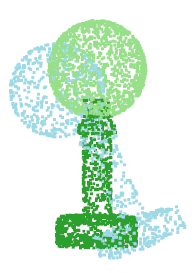} &
        \includegraphics[height=14mm]{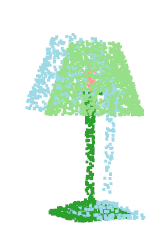} &
        \includegraphics[height=14mm]{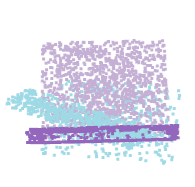} &
        \includegraphics[height=14mm]{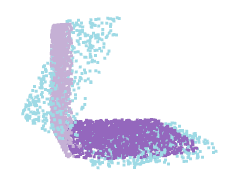} \\
        \parbox[t]{10mm}{\centering\rotatebox{90}{\scriptsize\textbf{PointCVaR}}} &
        \includegraphics[height=14mm]{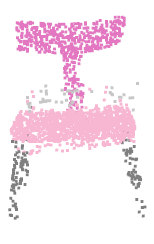} &
        \includegraphics[height=14mm]{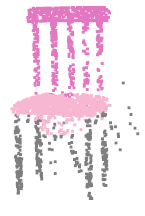} &
        \includegraphics[height=14mm]{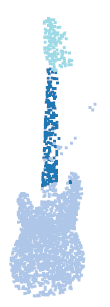} &
        \includegraphics[height=14mm]{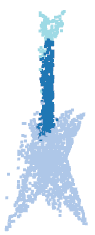} &
        \includegraphics[height=14mm]{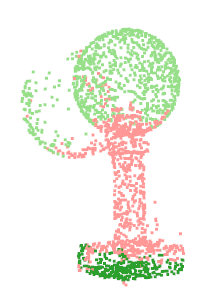} &
        \includegraphics[height=14mm]{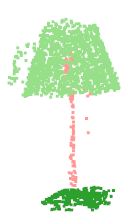} &
        \includegraphics[height=14mm]{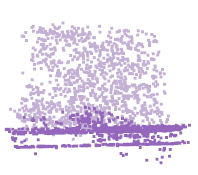} &
        \includegraphics[height=14mm]{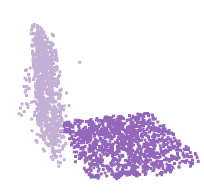} \\
        \parbox[t]{10mm}{\centering\rotatebox{90}{\scriptsize\textbf{GUARD}}}  &
        \includegraphics[height=14mm]{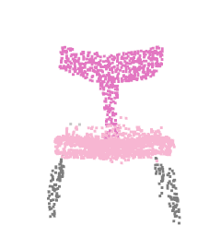} &
        \includegraphics[height=14mm]{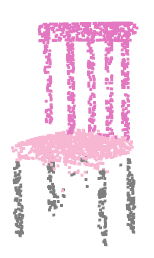} &
        \includegraphics[height=14mm]{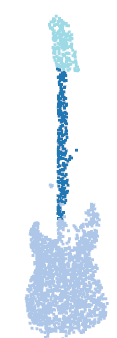} &
        \includegraphics[height=14mm]{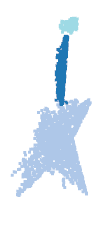} &
        \includegraphics[height=14mm]{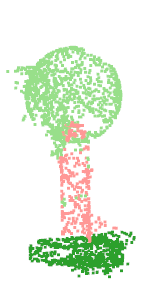} &
        \includegraphics[height=14mm]{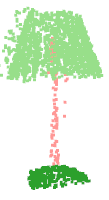} &
        \includegraphics[height=14mm]{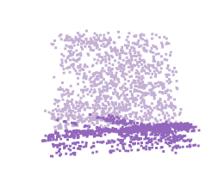} &
        \includegraphics[height=14mm]{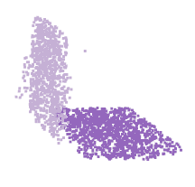} \\
    \end{tabular}%
}

    \caption{Visualization of segmentation results on representative ShapeNetPart categories under noisy conditions.}
    \label{fig:vis_results}
\end{figure*}
Similarly, Figure~\ref{fig:vis_results} shows representative categories from ShapeNetPart under synthetic noise.
Our method successfully removes ghost outliers and recovers clean, geometrically consistent part boundaries across diverse object categories.

\begin{figure*}[!t]
    \centering
    \subfloat[Scene 0019\label{fig:vis_scannet_0019}]{%
        \includegraphics[width=0.48\linewidth]{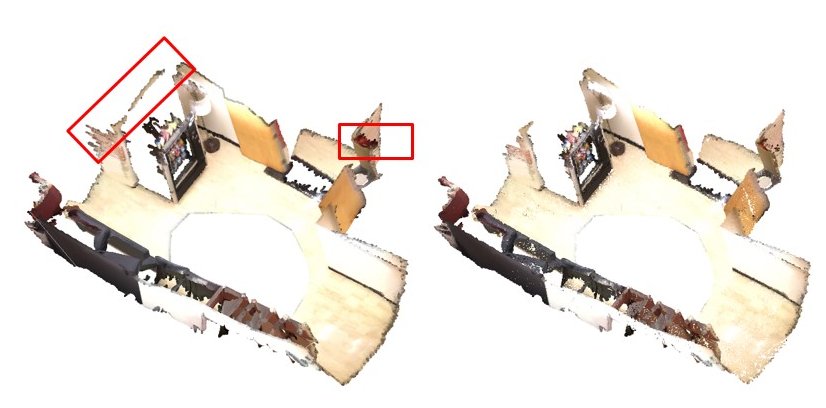}}
    \hfill
    \subfloat[Scene 0046\label{fig:vis_scannet_0046}]{%
        \includegraphics[width=0.48\linewidth]{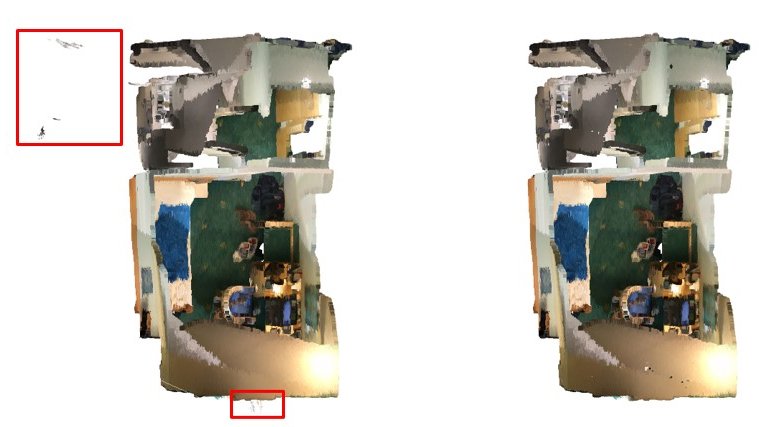}}
    \caption{
Qualitative results on ScanNet. Our method effectively removes floating ghost artifacts and global noise, yielding cleaner reconstructions in complex indoor scenes.
    }
    \label{fig:vis_scannet}
\end{figure*}

To further validate robustness in large-scale real-world scenes, we additionally evaluate our method on the ScanNet dataset, with qualitative results shown in Figure~\ref{fig:vis_scannet}.

\subsection{Ablation Experiment.}  

We ablate the uncertainty components to study their individual contributions. On ShapeNetPart, we compare using only the GP head (GGPH) versus only the entropy term.
As shown in Table~\ref{tab:ablation_shapenet}, GGPH provides stronger noise filtering (lower noise kept), while entropy alone attains slightly higher mIoU but tends to over-filter challenging yet informative regions, suggesting it is not a reliable proxy for geometric corruption.
Therefore, we use GGPH as the primary detector of structural inconsistencies, and entropy as a complementary refinement near locally ambiguous boundaries; their combination yields a better balance between completeness and reliability.

\begin{table}[!tb]
\centering
\caption{GGPH vs. entropy ablation on ShapeNetPart.}
\label{tab:ablation_shapenet}
\begin{tabular}{lcc}
\toprule
\textbf{Module} & \textbf{Avg. Noise Kept(\%)} $\downarrow$ & \textbf{Instance mIoU} $\uparrow$ \\
\midrule
Only Entropy & 16.1 & 0.7203 \\
Only GGPH    & 11.7 & 0.6672 \\
\bottomrule
\end{tabular}
\end{table}

On the real HDD dataset, we further ablate the components inside the uncertainty head and report outlier-detection precision/recall/F1 in Table~\ref{tab:ablation}.
Each module is complementary: MSN improves local robustness, TF enhances global reasoning, MdRFF captures multi-frequency structure, and SN stabilizes GP training.
Removing any component reduces F1, while the full model achieves the best accuracy on real scans.

\begin{table}[!tb]
\centering
\caption{Each component contributes to robustness.}
\label{tab:ablation}
\begin{tabular}{lccc}
\toprule
Method & Precision & Recall & F1 \\
\midrule
Baseline (RFF)   & 0.3358 & 0.2537 & 0.2890 \\
GUARD w/o MSN     & 0.9771 & 0.0953 & 0.1736 \\
GUARD w/o TF      & 0.7192 & 0.5979 & 0.6530 \\
GUARD w/o MdRFF   & 0.2851 & 0.8636 & 0.4287 \\
GUARD w/o SN      & 0.2108 & 0.1324 & 0.1626 \\
\midrule
GUARD (full)      & 0.8330 & 0.5897 & 0.6906 \\
\bottomrule
\end{tabular}
\end{table}

\subsection{Extended Analysis of Robustness and Generalization}
\subsubsection{Per-Category mIoU Results}
We report the per-category mIoU on ShapeNetPart to complement the main results in Table~\ref{tab:cat_miou}.
As shown in Table~\ref{tab:cat_miou}, our method achieves consistent improvements over the vanilla PointNet++ baseline across nearly all categories.
Notably, the performance gains are more pronounced on geometrically complex or fine-grained objects such as \textit{Cap}, \textit{Skateboard}, \textit{Chair}, and \textit{Knife}, indicating that our geometric uncertainty modeling enhances feature reliability for both smooth and intricate structures.
Meanwhile, categories with large planar or symmetric parts (e.g., \textit{Table}, \textit{Lamp}) also benefit from the hybrid denoising process, maintaining stable segmentation even under high corruption severity.
Overall, these per-category results further validate the robustness and generality of our approach across diverse object geometries.

\begin{table*}[!t]
\centering
\caption{Per-category mIoU results on ShapeNetPart(ghostcluster-s5) for different methods. 
We report results for PointNet++ and DGCNN backbones under three settings: Vanilla, PointCVaR, and our GUARD.}
\label{tab:cat_miou}
\resizebox{\textwidth}{!}{
\begin{tabular}{llc|cccccccccccccccc}
\toprule
\multirow{2}{*}{\textbf{Backbone}} 
& \multirow{2}{*}{\textbf{Method}} 
& \multirow{2}{*}{\textbf{Avg. mIoU} $\uparrow$} 
& \multicolumn{16}{c}{\textbf{Per-Category mIoU} $\uparrow$ on ShapeNetPart} \\
\cmidrule(lr){4-19}
& & 
& Earphone & Motorbike & Rocket & Car & Laptop & Cap & Skateboard & Mug 
& Guitar & Bag & Lamp & Table & Airplane & Pistol & Chair & Knife \\
\midrule
\multirow{3}{*}{PointNet++} 
& Vanilla    
& 0.6082
& 0.4786 & 0.5345 & 0.4609 & 0.5102 & 0.7431 & 0.5680 & 0.4366 & 0.7016 
& 0.7709 & \textbf{0.5754} & 0.7400 & 0.5774 & 0.6345 & 0.6380 & 0.6991 & 0.6624 \\

& +PointCVaR 
& 0.6049
& 0.5242 & 0.3726 & 0.4437 & 0.5739 & 0.7984 & 0.5998 & 0.4594 & 0.6300 
& 0.7889 & 0.4863 & 0.7091 & 0.6374 & 0.6656 & 0.6378 & 0.7057 & 0.6456 \\

& +\textbf{GUARD} 
& \textbf{0.7014}
& \textbf{0.6363} & \textbf{0.5841} & \textbf{0.4977} & \textbf{0.5836} & \textbf{0.8529} & \textbf{0.7008} & \textbf{0.6025} & \textbf{0.7126} 
& \textbf{0.8786} & 0.5645 & \textbf{0.7939} & \textbf{0.7246} & \textbf{0.7741} & \textbf{0.6660} & \textbf{0.8394} & \textbf{0.8108} \\
\midrule
\multirow{3}{*}{DGCNN} 
& Vanilla    
& 0.6200
& 0.5847 & 0.3857 & 0.4199 & \textbf{0.5224} & 0.7648 & 0.6289 & 0.4293 & 0.6433 
& 0.7915 & 0.6242 & 0.7600 & 0.6664 & 0.6700 & 0.5868 & 0.7725 & 0.6696 \\

& +PointCVaR 
& 0.6555
& 0.6501 & 0.4082 & 0.4530 & 0.5209 & 0.8413 & 0.6993 & 0.4469 & 0.5840 
& 0.8265 & 0.6831 & 0.7667 & 0.7124 & 0.7137 & 0.6307 & 0.8103 & 0.7409 \\

& +\textbf{GUARD} 
& \textbf{0.7320}
& \textbf{0.6647} & \textbf{0.4850} & \textbf{0.4900} & 0.4551 & \textbf{0.9074} & \textbf{0.7399} & \textbf{0.6646} & \textbf{0.8264} 
& \textbf{0.9186} & \textbf{0.7090} & \textbf{0.8363} & \textbf{0.8084} & \textbf{0.7591} & \textbf{0.7408} & \textbf{0.9096} & \textbf{0.7971} \\
\bottomrule
\end{tabular}}
\end{table*}

To further assess the efficiency and robustness of our uncertainty integration design, 
we compare the proposed \emph{Fusion} strategy with a two-stage \emph{Cascade} baseline.  
The \emph{Cascade} method sequentially applies two kinds of uncertainty filters, 
while the \emph{Fusion} approach combines them in a single step via a weighted sum, 
reducing computational overhead and improving stability.

\begin{table}[!tb]
\centering
\caption{Comparison between Fusion and Cascade strategies. 
mIoU measures segmentation accuracy, 
Noise Ratio (\%) denotes the proportion of residual noisy points, 
and mean latency (ms) reflects average inference time per sample.}
\label{tab:fusion_cascade}
\resizebox{\linewidth}{!}{
\begin{tabular}{lccc}
\toprule
\textbf{Method} & \textbf{mIoU} $\uparrow$ & \textbf{Noise Ratio (\%)} $\downarrow$ & \textbf{Mean Latency (ms)} $\downarrow$ \\
\midrule
Cascade & 0.6282 & 24.41 & 189.68 \\
\textbf{Fusion (ours)} & \textbf{0.7320} & \textbf{22.30} & \textbf{169.51} \\
\bottomrule
\end{tabular}
}
\end{table}

The results show that the \emph{Fusion} design improves segmentation performance by 
reducing the retained noise ratio and achieving a lower mean latency compared to the 
\emph{Cascade} counterpart, confirming its practical advantage for efficient denoising 
in robotic perception tasks.

\subsubsection{Additional Analysis on DGCNN on HDD Data}

During our experiments on the HDD dataset, we observe that the performance of 
\textbf{DGCNN + GUARD} is slightly lower than the vanilla DGCNN baseline, although it 
still significantly outperforms PointCVaR. To better understand this behavior, we 
conduct additional analysis focusing on the effect of the filtering ratio used in 
the denoising stage.

We find that the filtering ratio has a noticeable impact on the final segmentation 
performance. One possible explanation is that DGCNN relies heavily on local edge 
structures through the EdgeConv operator, which dynamically constructs graphs based 
on neighboring points. When too many points are filtered out, some informative 
points contributing to local graph construction may also be removed, potentially 
weakening the extracted features and degrading segmentation performance. In contrast, 
a more moderate filtering ratio preserves sufficient structural information while 
still removing noisy points.

To further investigate this effect, we evaluate several filtering ratios and report 
the results in Table~\ref{tab:dgcnn_ratio}. The results show that overly aggressive 
filtering leads to reduced performance, while a moderate filtering ratio achieves 
a better balance between noise removal and feature preservation.

\begin{table}[!tb]
\centering
\caption{Effect of filtering ratio for GUARD when applied to DGCNN on the HDD dataset. 
A moderate filtering ratio achieves the best balance between noise suppression and 
feature preservation.}
\label{tab:dgcnn_ratio}
\footnotesize
\setlength{\tabcolsep}{4pt}
\begin{tabular}{@{}c c c c c c c@{}}
\toprule
\shortstack{Filtering Ratio} 
& 0.70 & 0.75 & 0.80 & 0.85 & 0.90 & 0.95 \\
\midrule
mIoU $\uparrow$ 
& 0.7212 & 0.7295 & 0.7341 & 0.7373 
& \textbf{0.7425} & 0.7406 \\
\bottomrule
\end{tabular}
\end{table}

After analyzing the filtering ratio, we further study the effect of the 
fusion weight $\alpha$ that balances predictive entropy and geometric 
uncertainty $\sigma^2$ in the uncertainty estimation module. 
Recall that $\alpha$ controls the contribution of entropy versus 
geometry-conditioned uncertainty in the final uncertainty score.

The results are summarized in Table~\ref{tab:dgcnn_alpha}. 
Consistent with our earlier observations, smaller $\alpha$ values 
(which emphasize entropy) lead to slightly higher segmentation mIoU. 
However, this improvement mainly arises because entropy tends to 
remove ambiguous boundary points rather than true structural noise, 
resulting in a larger percentage of noisy points being retained. 
In contrast, larger $\alpha$ values rely more on geometry-conditioned 
uncertainty, which better captures structural inconsistencies and 
removes more fake points, though sometimes at the cost of a small 
reduction in mIoU.

\begin{table}[!tb]
\centering
\caption{Effect of the fusion weight $\alpha$ between predictive entropy 
and geometric uncertainty $\sigma^2$ on the HDD dataset using DGCNN. 
Lower $\alpha$ improves mIoU but retains more noisy points. 
The filtering ratio is fixed to 0.9.}
\label{tab:dgcnn_alpha}
\footnotesize
\setlength{\tabcolsep}{4pt}
\begin{tabular}{@{}c c c c c c@{}}
\toprule
$\alpha$ & 0.1 & 0.3 & 0.5 & 0.7 & 0.9 \\
\midrule
\shortstack{Noise Kept (\%) $\downarrow$}
& 18.8 & 13.5 & 9.9 & 7.8 & \textbf{6.7} \\
mIoU $\uparrow$ 
& \textbf{0.7834} & 0.7702 & 0.7571 & 0.7425 & 0.7301 \\
\bottomrule
\end{tabular}
\end{table}


\section{Discussion and Engineering Implications}

A key observation from the experiments is that geometric reliability and semantic confidence capture different aspects of point-cloud quality.
A corrupted structure may remain locally plausible and therefore receive a confident semantic prediction, even though its geometry is inconsistent with the underlying object.
The uncertainty analyses indicate that explicitly modeling geometric uncertainty provides complementary information to predictive entropy for identifying such measurements.
This distinction is particularly important for structured reconstruction artifacts, where semantic confidence alone may not adequately reflect the reliability of the observed geometry.

The results also highlight a tradeoff between artifact removal and preservation of information useful for segmentation.
More aggressive filtering can suppress additional corrupted measurements, but may also remove informative points that contribute to part recognition.
Conversely, retaining more points may preserve discriminative geometric details while allowing some unreliable measurements to remain.
Therefore, segmentation accuracy alone does not fully characterize denoising quality.
For engineering perception, both the reliability of the retained geometry and its usefulness for identifying components should be considered when selecting the filtering strategy.

For robotic disassembly, GUARD serves as a perception-stage mechanism for converting imperfect 3D measurements into more reliable part-level information.
By associating geometric representations with uncertainty and combining this information with semantic predictions, the framework can reduce the influence of duplicated or inconsistent structures before the resulting representation is used by downstream manipulation or planning modules.
The present experiments evaluate this capability through point filtering, part segmentation, and uncertainty-based corruption detection across the HDD, ShapeNetPart, and ScanNet datasets.
The present study focuses on perception-level reliability, while future work will examine how improvements in geometric reliability translate into downstream robotic manipulation and disassembly performance.

\section{Conclusion}
This study presented GUARD, a geometric uncertainty-aware framework for unreliable-point filtering and segmentation of engineering point clouds. By combining multi-scale geometric representations, Gaussian Process-based uncertainty estimation, and semantic entropy, GUARD distinguishes geometric reliability from confidence in part identification within a single-pass architecture. Evaluation on real HDD scans demonstrates improved PointNet++ part segmentation, while complementary benchmarks and ablations characterize corrupted-point detection, computational efficiency, and backbone-dependent filtering behavior. The results show that artifact suppression and preservation of informative geometry must be considered jointly rather than treating segmentation accuracy as the sole indicator of perception quality. GUARD therefore contributes both a computational method and empirical understanding of reliability-aware 3D interpretation, providing a perception-level foundation for component handling and robotic disassembly under imperfect sensing conditions.

%
%
\bibliographystyle{IEEEtran}
\bibliography{main}
\end{document}